\documentclass[lettersize,journal]{IEEEtran}
\usepackage{array}
\usepackage[caption=false,font=normalsize,labelfont=sf,textfont=sf]{subfig}
\usepackage{textcomp}
\usepackage{stfloats}
\usepackage{multirow}

\usepackage{verbatim}
\usepackage[cmex10]{amsmath}
\usepackage{amssymb}
\usepackage{amsfonts}
\usepackage{amsthm}
\usepackage[ruled,vlined]{algorithm2e}
\usepackage{algpseudocode}
\usepackage{graphicx}
\usepackage[inkscapelatex=false]{svg}
\usepackage{epsfig}
\usepackage{cite}
\usepackage{picinpar}
\usepackage{tensor}
\usepackage{color}
\usepackage{booktabs}
\usepackage{siunitx}
\usepackage{soul} 
\usepackage{url}
\usepackage[pdfa]{hyperref}
\hypersetup{
    colorlinks=true,
    linkcolor=blue,
    filecolor=magenta,      
    urlcolor=blue,
    }

\usepackage{flushend}
\usepackage{colortbl}
\usepackage{pifont}
\usepackage{alltt}
\usepackage{epstopdf}
\usepackage{pbox}

\graphicspath{{figures/}}
\DeclareGraphicsExtensions{.eps}
\theoremstyle{definition}

\makeatletter
\DeclareFontFamily{U}{tipa}{}
\DeclareFontShape{U}{tipa}{m}{n}{<->tipa10}{}
\newcommand{\arc@char}{{\usefont{U}{tipa}{m}{n}\symbol{62}}}%

\newcommand{\arc}[1]{\mathpalette\arc@arc{#1}}

\newcommand{\arc@arc}[2]{%
  \sbox0{$\m@th#1#2$}%
  \vbox{
    \hbox{\resizebox{\wd0}{\height}{\arc@char}}
    \nointerlineskip
    \box0
  }%
}

\makeatother

\usepackage{siunitx}
\DeclareSIUnit\mt{\milli\tesla} 
\usepackage{gensymb}

\begin{document}

\title{
Iterative Shaping of Multi-Particle Aggregates \\based on Action Trees and VLM
}

\author{Hoi-Yin Lee, \IEEEmembership{Member,~IEEE}, Peng Zhou,~\IEEEmembership{Member,~IEEE},  Anqing Duan,~\IEEEmembership{Member,~IEEE},\\ Chenguang Yang,~\IEEEmembership{Fellow,~IEEE}, and David Navarro-Alarcon,~\IEEEmembership{Senior Member,~IEEE}%
\thanks{This work is supported by the Research Grants Council of Hong Kong under grant number 15201824. \textit{Corresponding author: David Navarro-Alarcon.}}%
\thanks{H.-Y. Lee and D. Navarro-Alarcon are with the Department of Mechanical Engineering, The Hong Kong Polytechnic University, KLN, Hong Kong. hyinlee@polyu.edu.hk, dnavar@polyu.edu.hk}
\thanks{P. Zhou is with the School of Advanced Engineering, The Great Bay University, Dongguan, China. peng.zhou@ieee.org}
\thanks{A. Duan is with the Department of Robotics, Mohamed Bin Zayed University of Artificial Intelligence, UAE. anqing.duan@mbzuai.ac.ae}
\thanks{C. Yang is with the Department of Computer Science, University of Liverpool, Liverpool, UK. cyang@ieee.org}
}

\markboth{Iterative Shaping of Multi-Particle Aggregates based on Action Trees and VLM}{}

\maketitle

\begin{abstract}
In this paper, we address the problem of manipulating multi-particle aggregates using a bimanual robotic system. 
Our approach enables the autonomous transport of dispersed particles through a series of shaping and pushing actions using robotically-controlled tools.
Achieving this advanced manipulation capability presents two key challenges: high-level task planning and trajectory execution. 
For task planning, we leverage Vision Language Models (VLMs) to enable primitive actions such as tool affordance grasping and non-prehensile particle pushing. 
For trajectory execution, we represent the evolving particle aggregate's contour using truncated Fourier series, providing efficient parametrization of its closed shape. 
We adaptively compute trajectory waypoints based on group cohesion and the geometric centroid of the aggregate, accounting for its spatial distribution and collective motion.
Through real-world experiments, we demonstrate the effectiveness of our methodology in actively shaping and manipulating multi-particle aggregates while maintaining high system cohesion.

\end{abstract}

\begin{IEEEkeywords}
Robot manipulation, shape control, action trees, multi-particle manipulation, VLM.
\end{IEEEkeywords}

\section{Introduction}
\IEEEPARstart{T}{hroughout} history, the task of guiding large sheep flocks to designated pastures has relied primarily on herding dogs such as Border Collie. 
Rather than directly pushing individual sheep, which risks scattering the flock, these dogs skillfully maneuver around the group, guiding the entire flock as a \emph{cohesive} unit towards the desired location \cite{lien2004shepherding}.
This herding technique is not limited to the animal world, it is also commonly observed in our human daily lives as we often adopt similar strategies when dealing with dispersed elements. 
Taking floor sweeping as an example. 
Instead of painstakingly collecting each speck of dust individually, we gather them into a cohesive pile before sweeping it away with a dustpan. 
This intuitive approach mirrors the principles of herding, where the focus is on managing the group as a whole, rather than individually controlling each element.

This manipulation method is well-suited for multi-particle aggregation tasks, which involve consolidating dispersed elements and guiding them towards a target location (e.g., a storage area) while maintaining the group's cohesion. 
Taking inspiration from these examples, we can develop bio-inspired herding-like approaches for robots to automatically gather, shape, and transport multi-particle aggregates.

During the initial consolidation stage, this type of robotic herder can skillfully maneuver around the dispersed elements, gradually changing the overall shape and guiding it towards a target point. 
This strategy ensures that individual objects are not simply pushed in isolation, but rather gathered into a cohesive unit. 
Throughout this process, the robot can compute and adjust the shape and cohesion of the ensemble, thus, maintaining its compactness and preventing fragmentation.

With the elements gathered into a cohesive pile, the robotic herder can then focus on shaping and guiding group towards the designated location (or even along a path). 
By treating the ensemble as a unit, the robot can actively change the group's morphology to effectively guide it through obstacles and tight spaces. 
This process can also reduce the risk of losing individual components along the way, as it ensures that the particle ensemble remains cohesive throughout the task.

\begin{figure}[t]
    \centering
    \includegraphics[width=1\linewidth]{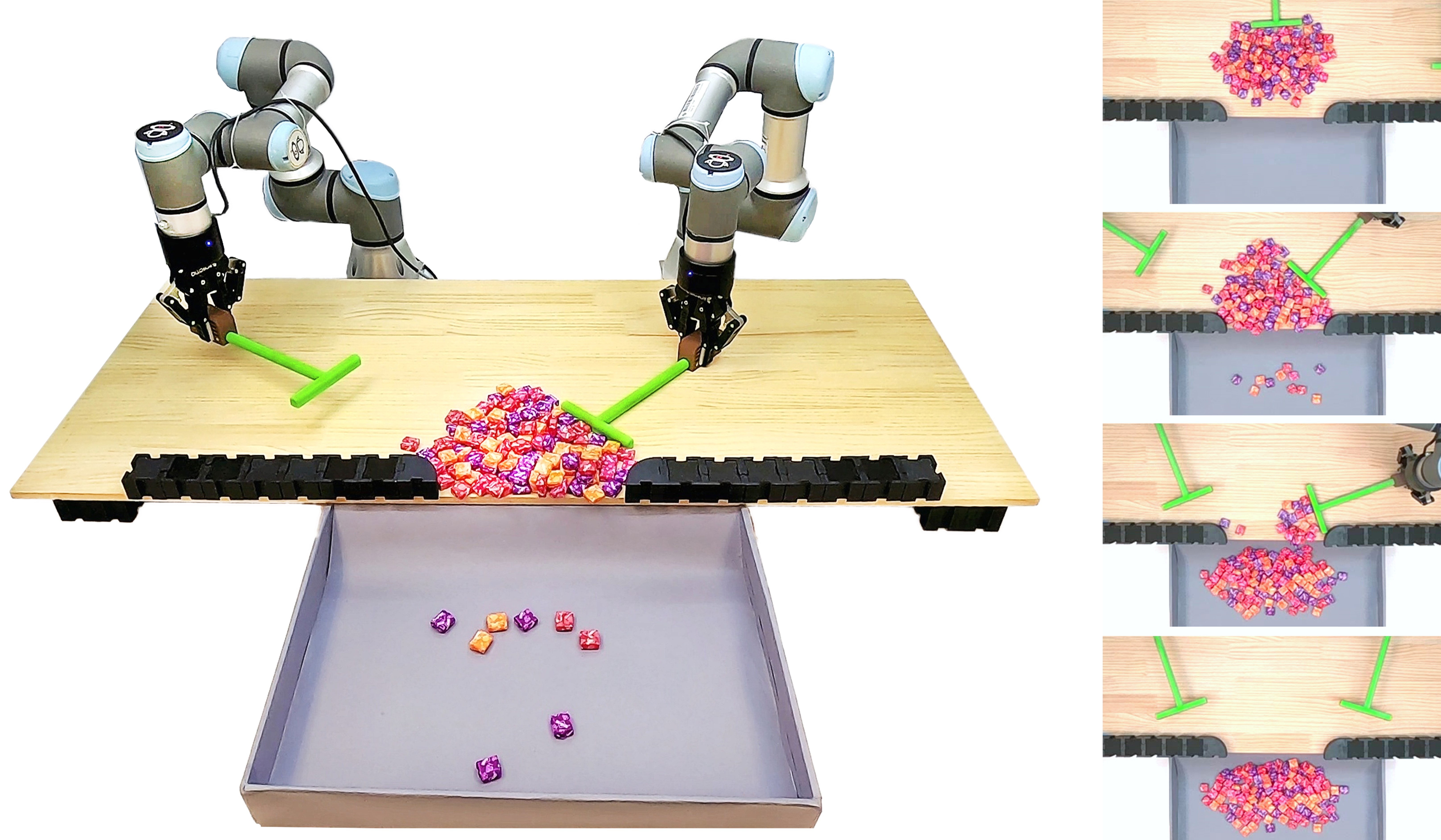}
    \caption{Setup of the addressed multi-particle shaping task.}
    \label{fig:setup}
\end{figure}

\subsection{Related Work}
To mimic this herding-like guiding strategy, two main approaches have been explored in the literature: multi-robot systems \cite{zhang2024distributed, zhang2024heterogeneous, vardy2019landmark, zhang2022collecting, van2024reactive, hamandi2024robotic} and robotic manipulators \cite{wang2023dynamic, zimmermann2014automated, suh2021surprising, schenck2017learning, tuomainen2022manipulation, lu2021excavation, takahashi2021uncertainty, cherubini2020model, zhu2019data, matl2020inferring}.
For the former, researchers have developed different active methods, for example, \cite{zhang2024distributed} presented a distributed strategy to herd groups of active (robotic) evaders to a predefined area, akin to a sheepdog pushing the outmost evader. 
Similarly, \cite{zhang2024heterogeneous} developed an adaptive density-based interaction controller for trapping heterogeneous  targets with swarm robots, allowing them to self-organize and adjust the encirclement based on target strength. 
Dynamic path planning is also crucial for herding groups to a destination while avoiding obstacles, as demonstrated in \cite{van2024reactive} and \cite{hamandi2024robotic}.
Some multi-agent works have also addressed the shaping of ensembles, see e.g., \cite{vardy2019landmark} and \cite{strombom2018robot}.

\begin{figure*}[t]
    \centering
    \includegraphics[width=1\linewidth]{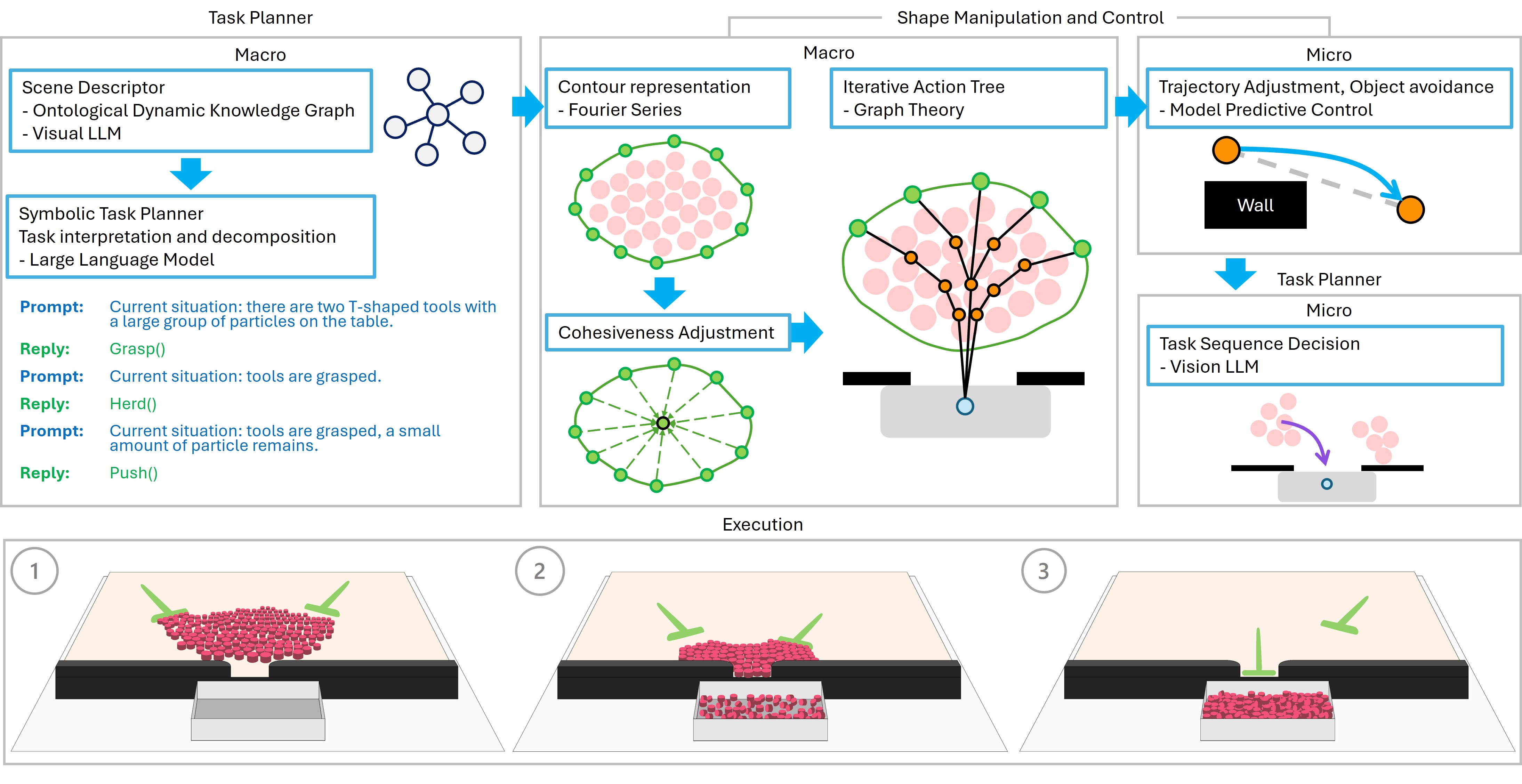}
    \caption{System Structure: The system consists of a Task Planner with a Shape Manipulation and Control module. An ontological knowledge graph and VLM describe the scene, interpreted by an LLM for task decomposition. Shape is represented by Fourier Series to generate an action tree based on particle cohesiveness. An MPC guides manipulation while avoiding obstacles, with the VLM confirming particle status if detection becomes difficult.}
    \label{fig:flowchart}
\end{figure*}

Robotic manipulators have also been used for shaping and manipulating multi-particle aggregates.
For example, \cite{suh2021surprising, tuomainen2022manipulation, lu2021excavation, schenck2017learning} have proposed learning-based methods to compute the dynamics of granular pieces and thus perform the task.
To understand the scene and compute optimal actions, researchers have developed efficient strategies to learn models from visual data and text-based scene interpretation \cite{lin2022learning,wu2023daydreamer,finn2017deep,shi2024robocraft,liu2024robouniview,kawaharazuka2024continuous,wake2024gpt,huang2023visual,mei2024replanvlm,hylee2024nonprehen,jeong2024survey,wang2024large,gao2024physically,zeng2023large}.

While there has been substantial work in multi-object/particle manipulation, the preservation of group cohesion during the manipulation process remains an underexplored area. 
Adopting shepherding strategies based on iterative shaping of the aggregate is a promising direction that warrants further investigation.

\subsection{Problem Formulation}
Our objective in this work is to develop a tool-based method to ``guide" a group of particles that are initially located outside a designated gate area. 
These particles must be iteratively pushed through a gate to reach a containment box. 
This can be viewed as a herding problem, where the particles are akin to a flock of sheep (although passive) and the robotic tool serves as a ``sheepdog" to guide them towards the desired goal location. 
A key requirement to preserve cohesion among the group of particles throughout the manipulation task.
In simple words, we want the aggregate to \textbf{Stay Together and Stay Connected}.

In our proposed method, the guiding and incremental pushing of the ensemble is referred to as ``herding". 
The slot or space through which the particles must pass to enter the container is called the ``gate".
The term ``cohesion" is used to describe the degree to which the group of particles is united, compact, and stuck together. 
It refers to how tightly packed or cohesive the particle group is as a whole.

\subsection{Contributions}
The original contributions of this work are as follows:
\begin{itemize}
\item A novel contour-based strategy to shape and transport multi-particle systems with non-prehensile actions.
\item An iterative action tree for path planning that preserves the cohesion and holistic nature of the particle group.
\item A cohesiveness metric to quantify the compactness of the particle ensemble.
\item A LLM-based planner that leverages the feedback shape to guide the herding of particles.
\end{itemize}

The rest of this manuscript is organized as follows: Sec. \ref{sec:method} presents the detailed methodology of our approach, Sec. \ref{sec:results} reports the empirical results and evaluations, and Sec. \ref{sec:conclusions} provides the final conclusions drawn from this work.

\section{Methodology} \label{sec:method}
To manipulate a group of particles, a systemic methodology has been developed, as depicted in Fig. \ref{fig:flowchart}. The entire process is divided into two aspects: (1) shape manipulation and control, and (2) task planning. Each aspect has a macro and a micro section, where the macro provides an approximate solution, which is then refined by the micro part.

\subsection{Macro-scale Shape Manipulation} 
\subsubsection{Shape Representation}\label{sec:fourier}
Representing and controlling the shape of a particle group poses unique challenges compared to manipulating a deformable soft object. Unlike a soft object, whose shape is bounded by its volume and mass distribution, the configuration of a particle group can be highly unconstrained. The particles may be distributed in an arbitrary spatial arrangement, with no inherent limits on their relative positions or the overall shape of the ensemble.
A small movement of a single particle may not contribute to the majority of shape-changing. For example, the particles can squeeze into the spaces among the ensemble with no significant total shape deformation. Thus, instead of an accurate model to represent the details ensemble's shape and the dynamic motion of particles, we propose an arbitrary shape representation approach based on Fourier descriptors, see \ref{fig:flowchart}. By representing the boundary of the particle group using a compact set of Fourier coefficients, we can track and mould the macro-scale shape without requiring a detailed dynamical model of the individual particle motions.

Since no prior information is available about the shape of the particle group, we extract the contour with computer vision and represent the contour with a Fourier series. The complex-valued function $f(\tau)$ is used to express the contour as:
\begin{equation}
f(\tau) = \sum_{n = 0}^{N} c_n e^{i n \frac{2\pi \tau}{\rho}}
\end{equation}
where $c_n$ are the coefficients of the Fourier series, $\rho = 2 \pi$ is the period of the function, and $N$ is the number of harmonics.
The coefficients $c_n$ can be computed with the integral:
\begin{equation}
c_n = \frac{1}{2\pi} \int_0^{2\pi} f(\tau) e^{-i n \tau} d\tau
\end{equation}
and the contour is reconstructed by summing the contributions of the Fourier coefficients at each time point $\tau$: 
\begin{equation}
\mathbf{z}(\tau) = \sum_{n = 0}^N c_n e^{i n \frac{2\pi \tau}{P}}
\end{equation}
where $\mathbf{z}(\tau) = x(\tau) + i y(\tau)$ is the complex-valued position of the contour at time $\tau$ with the range set to $[0, 2\pi]$. The scalar $N$ represents the number of \emph{finite} Fourier coefficients used.

The choice of the number of Fourier harmonics used to represent the particle group's shape is a key parameter in this approach. A larger number of harmonics allows for a more detailed and accurate shape representation, capturing finer contour features of the particle distribution. However, this increased complexity also comes with more redundant contour points and a rougher, less streamlined overall shape.
Conversely, using a smaller number of harmonics results in a more simplified, smoother contour representation, but at the cost of losing some of the detailed shape information. For the purposes of this particle herding task, we are primarily interested in controlling the macro-scale shape of the particle ensemble, rather than tracking its micro-scale features. Therefore, a lower number of harmonics is preferred, as it provides a sufficiently accurate yet compact shape representation that can be effectively servoed by the robot.

\subsubsection{Cohesiveness Metric}
Maintaining the cohesiveness of a particle group and keeping the particles compact are crucial objectives. To quantify this cohesiveness, we introduce a cohesiveness metric that measures how tightly the particles are encircled within the group's contour.
Based on the geometric properties of the particle group, we observe that the distance between any point on a circle and its centroid is always constant. In contrast, the distance between a corner point of a rectangle/square and its centroid is not identical to any other non-corner point, as it is much farther away. Therefore, a regular circle represents the optimal enclosure for a group of objects given the same density, and the size of the group is irrelevant in this context. The same principle can be applied to other shapes, where a square is better than a wide rectangle in terms of cohesiveness.

We refer to this concept as the shape regularity, which is calculated based on the average distance between a set of $n$ points on the contour of the particle group and the group's centroid, compared to the optimal, minimalistic area. The shape regularity metric ranges from 0 to 1, where higher values indicate a more regular, compact shape.

Density is another important factor in determining the level of compactness within a region. It can be calculated by finding the ratio between the area occupied by the particles and the total area of the particle group.
We can mathematically express the cohesiveness metric as follows:
\begin{equation} \label{eq:cohesion}
\zeta = \frac{\sqrt{\frac{\alpha}{\pi}}}{\frac{1}{n}\sum_{i=1}^n ||\mathbf{p}_i - \text{mean}(\mathbf p)||} \times \frac{\alpha}{\beta} \times 100\%
\end{equation}
where $\alpha$ is the area occupied by the particles and $\beta$ is the total area of the particle group calculated by Fourier-series analysis.
$\mathbf{p}_i$ represents the $i$-th point on the Fourier-based contour of the particle group.
The function $\text{mean}(\mathbf p)$ is the centroid of the particle group.
The first part of the equation measures the shape regularity of the particle group by comparing it with the optimal encirclement (i.e. a circle), where the radius $r$ is solved from the area equation $\alpha = \pi r^2$ and is divided by the average Euclidean distance of the contour points from the centroid. The second part calculates the density of the particle group by dividing the occupied area of the particles $\alpha$ by the total area of the whole group $\beta$.

By combining these two factors, the cohesiveness metric $\zeta$ is obtained, which ranges from 0 to 100\%, with higher values indicating a more cohesive particle group. This metric provides a quantitative measure of the particle group's cohesion, which can be used to optimize the particle arrangement and overall system performance.

\begin{figure}[t]
    \centering
    \includegraphics[width=1\linewidth]{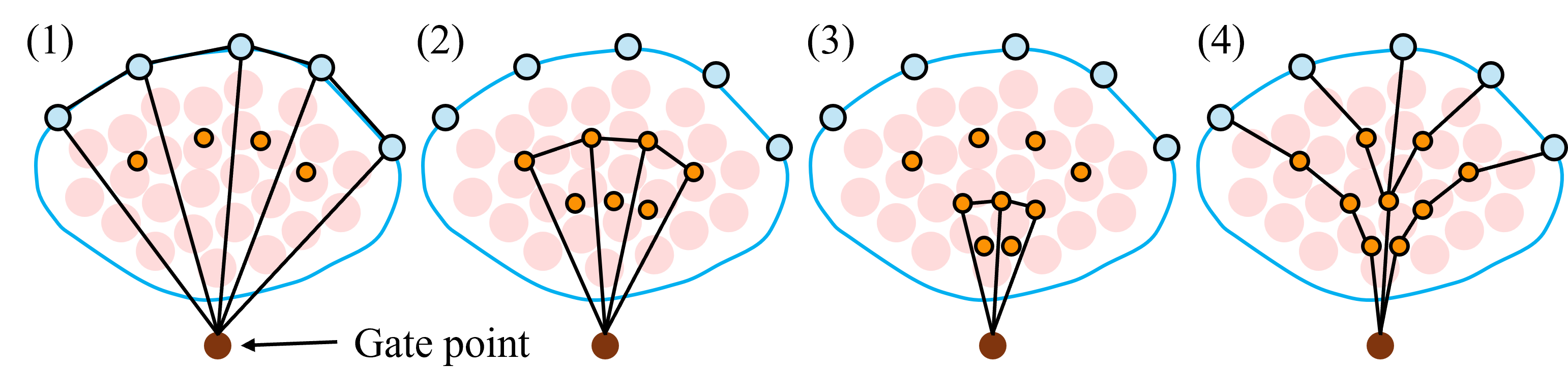}
    \caption{Conceptual representation of the proposed iterative action tree. (1)--(3): The orange point represents a waypoint on a trajectory, which is positioned at the centroid of a triangle. (4): The overall trajectory can be obtained by applying graph theory principles to connect these strategically placed waypoints.}
    \label{fig:treeStructure}
\end{figure}

\begin{algorithm}[t]
\caption{Iterative Action Tree for Path Planning}\label{alg}
\KwIn{$\mathbf{P}$, $\text{gate}$ \Comment{Set of farthest points, gate location}}
\KwOut{$\mathbf{\Pi}$ \Comment{Set of trajectories}}
$\mathbf{C}_0 \gets \text{ComputeInitialCentroids}(\mathbf{P}, \text{gate})$; \\
$\mathbf{C} \gets \mathbf{C}_0$; \\ 
$i \gets 1$;\\
\While{$|\mathbf{C}_i| > 2$}{
    \For{$j \gets 1$ to $(|\mathbf{C}_i|-1)$}
    {
        $\mathbf{c} \gets \text{FindCentroid}(\mathbf{C}_{i-1,j}, \mathbf{C}_{i-1,j+1}, \text{gate})$;\\
        $\mathbf{C}_i \gets \mathbf{C}_i \cup \mathbf{c}$;\\
    }
    $i \gets i + 1$; \\
    $\mathbf{C} \gets \mathbf{C} \cup \mathbf{C}_i$;\\
}
$\mathbf{G} \gets \text{Graph}(\text{gate}, \mathbf{C}, \mathbf{P})$; \\
$\mathbf{\Pi} \gets \text{ApplyDijkstra}(\mathbf{G})$; \\
\KwRet{$\mathbf{\Pi}$}
\end{algorithm}

\subsubsection{Path Planning: Iterative Action Tree} \label{sec:trajPlanning}
To guide the particle group towards the designated gate location, we propose a path-planning approach based on an iterative action tree. 
This method leverages the spatial distribution of the particles to define a sequence of waypoints that the robotic tool can follow to effectively herd the ensemble.

In the first step of our approach, we identify a subset of particles that are farther away. These distant particles represent the most outlying members of the aggregated group that require guidance to be brought back towards the target.
We utilize the cohesion metric to analyze which specific particles are contributing to a lower overall cohesion of the group. If the cohesion is observed to be low, we select the particles that are farther away from the group's centroid for targeted manipulation. Conversely, if the cohesion is relatively high, we focus our attention on the particles that are farthest from the goal location. This strategic selection of particles allows us to efficiently improve the compactness of the aggregated group while also drawing it back towards the desired target.
The distance between the points taken is less than the length of the tool segment to avoid losing `sheep' during the manipulation stage.
In our case, we set the number of points to be 5 based on our tool's dimension and we refer to these points as $\mathbf{P}$.

From this set of 5 farthest particles, we compute the initial centroids by finding the centroids of the triangles formed by connecting each pair of points with the gate location, see Fig. \ref{fig:treeStructure} and Alg. \ref{alg}.
Starting from these initial centroids, we then iteratively refine the trajectory by computing new centroids. This is done by finding the centroid of the triangle formed by the two consecutive centroids from the previous iteration and the gate location. In other words, the new centroid is calculated as the average of the two previous centroids and the gate point (see Fig. \ref{fig:treeStructure}). This iterative process effectively shrinks the size of the particle group with each iteration, as the centroids move closer to the gate. The algorithm continues this process until only two centroids remain, resulting in a sequence of centroid sets that progressively reduce the spatial extent of the particle distribution.

Next, we apply graph theory to the path-finding. We treat the gate location as the root of a graph, and the computed centroids with the farthest points as the graph nodes. To identify the optimal trajectories from the farthest particles to the gate, we apply a path-finding algorithm: Dijkstra's algorithm. This allows us to construct a set of candidate trajectories that the robotic tool can follow to herd the particles back towards the desired goal location.

\begin{figure}[t]
    \centering
    \includegraphics[width=1\linewidth]{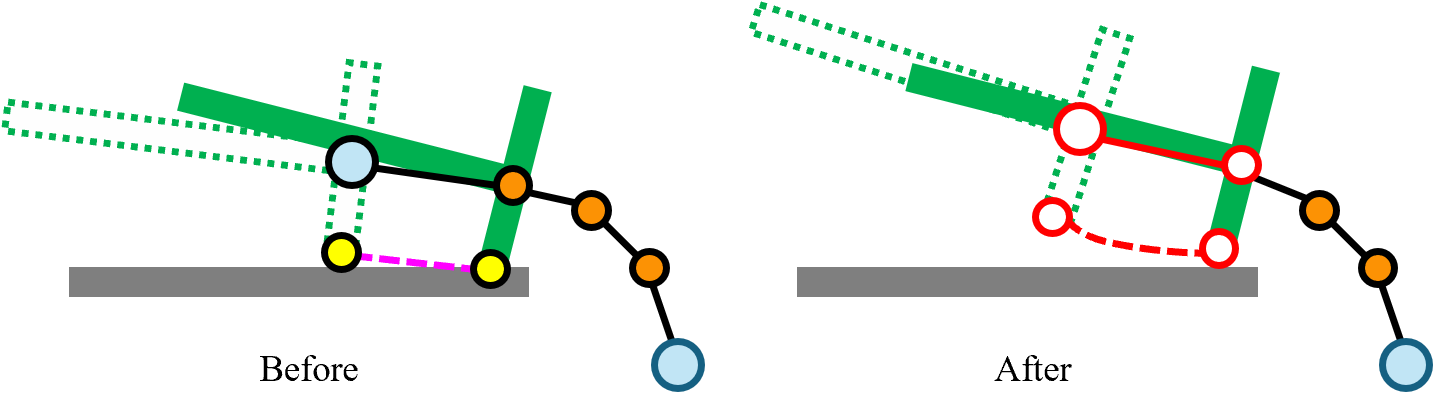}
    \caption{Before the micro-scale shape refinement, the tooltip (the purple dashed trajectory) will hit the grey wall. After the refinement, the tooltip (the red dashed trajectory) will be smoothly avoiding the grey wall.}
    \label{fig:mpc}
\end{figure}

\subsection{Micro-scale Shape Manipulation} \label{sec:mpc}
In each iteration, the robotic tool approaches and moves along the waypoints of a candidate trajectory. To mimic the herding behaviour where the group is gradually moved closer to the gate, the tool only travels a short distance in each herding task. Specifically, the tool moves between two waypoints at a time, unless the final waypoint is not within the ensemble or the remaining group size is too small and close to the gate.

After generating the candidate trajectories using the centroid-based approach, we apply an off the shelf model predictive control (MPC) framework to refine the trajectories and ensure collision-free motion of the robotic tool. The goal of this refinement is to optimize the tooltip trajectory to avoid obstacles, such as the grey wall shown in Fig. \ref{fig:mpc}.

We first obtain the dimensions of the tool from the computer vision system, which allows us to determine the originally planned start and end points for the tooltip (shown as yellow points in Fig. \ref{fig:mpc}). We then apply a refinement controller using these initial tooltip waypoints as the optimization objective. The controller computes the refined, obstacle-avoiding trajectory (shown as the red dashed line in Fig. \ref{fig:mpc}) by minimizing a cost function that penalizes deviations from the reference trajectory and control effort while satisfying the dynamic constraints and obstacle avoidance requirements.

The refinement controller uses a dynamic model of the tool and the environment to predict the future states of the system. This model includes the differential kinematic equations of the tool, as well as the positions and dimensions of the obstacle in the workspace, such as the gate walls.

The optimization problem is formulated to minimize a cost function that penalizes deviations from the desired trajectory, as well as control inputs that exceed the robot's actuation limits (e.g. having a sudden quick move). The optimization is subject to constraints that ensure the robot's predicted states do not collide with any obstacles over the prediction horizon.
The controller optimization problem is formulated as:
\begin{equation}
J = \text{arg}\min\limits_{\mathbf u} \ \sum_{i=0}^{H-1} ||\boldsymbol \chi_i - \boldsymbol \chi_\text{ref}||_Q^2 + ||\mathbf u_i||_R^2
\end{equation}
where it is subject to $\dot{\boldsymbol \chi} = f(\boldsymbol \chi, \mathbf{u})$ and $g(\boldsymbol \chi_i) \geq 0$.
$\boldsymbol \chi = [x, y, \theta]^\top$ is the state vector of the x-y coordinates and orientation of the tooltip, $\mathbf u = [v, \omega]^\top$ is the control input vector represents the linear velocity $v$ and angular velocity $\omega$ of the tooltip, $H$ is the prediction horizon, $\boldsymbol \chi_\text{ref}$ is the constant reference state, and $Q$ and $R$ are positive definite weight matrices. The function $g$ ensures that the tooltip maintains a minimum distance from the obstacles (i.e. the gate wall).

By solving the optimization problem, we can refine the trajectory of the tool and determine the optimized boundary for the tooltip's motion.

\subsection{High-Level Symbolic Task Planner} \label{llm}
To conduct a long-horizon task that involves logical action planning, we implemented an LLM-based planner for high-level symbolic task decomposition and decision-making. The motion for aggregating particles can be spit into macro and micro planning, see Fig. \ref{fig:flowchart}. In terms of the macro aspect, it involves task interpretation and action control and we use a description-based Task Planner for this. In terms of the micro aspect, it involves the detail decision-making for small adjustments to improve the performance. For this, we implement a vision-based task planner.

\subsubsection{Description-based Task Planner}
To interpret the scene information and the task requirement to generate the next action with predefined action functions, we implement a high-level symbolic planner similar to our previous studies \cite{lee2023distributed, hylee2024nonprehen}.
Instead of fine-tuning an LLM, we use the prompt approach for this herding task. 

We leverage the ontological dynamic knowledge graph (ODKG) presented in \cite{lee2023distributed} to store and provide the existing physical and virtual interaction information about the scene. We then apply the VLM to update the ODKG with the latest visual observations, generating textual grounding information that represents the current state of the environment.
Then, an LLM takes the natural language input along with the grounding information as input and outputs the appropriate next action function. This iterative process continues, with the system executing the action, observing the updated scene, and generating the next action based on the evolving grounding information.

The available action functions are adjusted to fit our current particle aggregation scenarios, including ``\texttt{grasp}, \texttt{herd}, \texttt{push}, \texttt{check}, \texttt{release}". The detailed description of the functions is stated in Table \ref{table:tbl_motionFunctions}.

To increase robustness, a few-shot learning method is adopted, and several concrete examples are given in the prompt to illustrate the details of the predefined actions. 
Consider a simple particle aggregation task, the possible action sequence could be ``\texttt{grasp()}, \texttt{herd()}, \texttt{check()}, \texttt{push()}, \texttt{check()}, \texttt{release()}", where the dual-arm robot grasps tools, then applied the shape manipulation module to compute the trajectory, and the dual-arm robot herds the `sheep' back to the desired point. The system checks to confirm if all the `sheep' are inside the gate, and the tools are returned to their original places when the task is considered completed by the micro task planner (more details are mentioned in the next section).

\begin{table}[t]
\centering
\caption{Motion Functions and Descriptions}
\label{table:tbl_motionFunctions} 
\begin{tabular}{p{0.35\linewidth} | p{0.55\linewidth}} \toprule
Motion Functions & Descriptions \\ \midrule
\tt{Grasp()} & To move and grasp the tool. \\ 
\tt{Herd()} & To slightly push the particle closer to the gate. \\ 
\tt{Push()} &  To push the particle directly to the gate. \\
\tt{Check()} &  To check the current herding progress and calculate the trajectories for the next round if needed.\\
\tt{Release()} &  To release the tool back to its original location and the robot to its home position.\\\bottomrule
\end{tabular}
\end{table}

\subsubsection{Vision-based Task Planner}
While the text description-based task planner is sufficient for simple, fast action control, it may miss some small details that require further adjustments to enhance performance. For instance, when only a small amount of particles remain, basic image processing could fail to detect them due to factors like varying lighting conditions affecting color detection accuracy.

To ensure no particles are left behind and the task is truly completed, we complement the image processing with a vision language model (VLM). When the image processing suggests the particle count is low, the system captures the scene and passes it to the VLM for verification.
This verification process is triggered by the ``check()" action command from the description-based task planner or the group size of the particle is smaller than a certain size. 
If there are no remains, the task is considered complete, and the ``release()" command is sent to the robot, instructing it to return the tools and move back to the home position.
If the VLM detects any remaining particles, the system continues the task. 

As the particle ensemble becomes smaller (i.e., with only a few particles left), some of the candidate trajectories generated in Sec. \ref{sec:trajPlanning} may become redundant and unnecessary to follow.
To address this, the VLM takes the candidate trajectories along with the current image as input, and outputs the most suitable starting point for the pushing action. The system then matches this starting point to the available candidate trajectories and selects the corresponding path to follow.

\begin{figure*}[t]
    \centering
    \includegraphics[width=1\linewidth]{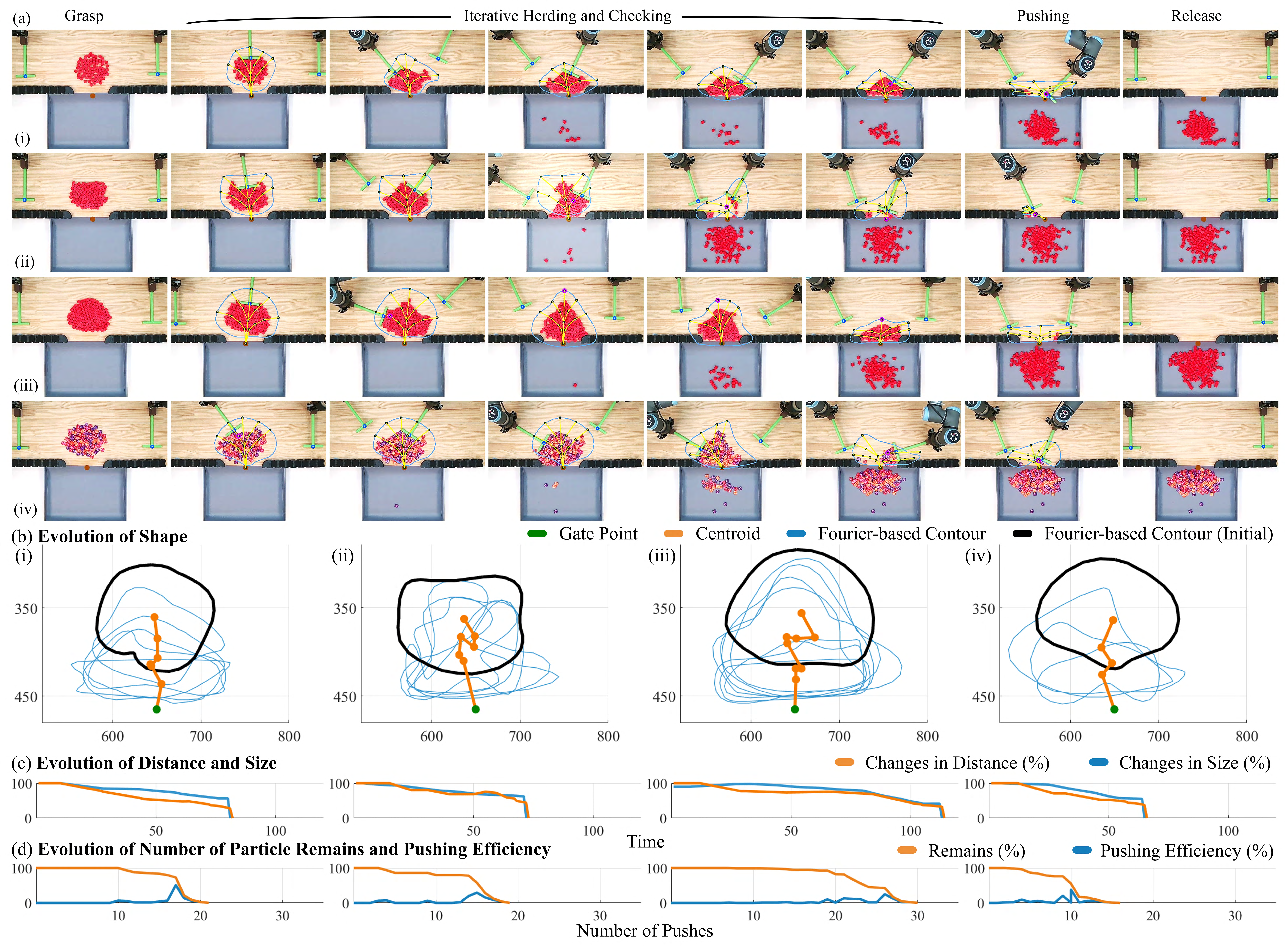}
    \caption{Experiments: (a) Different shapes and amounts of particles are used to evaluate the performance of the proposed shape representation method. The contour of the particle is expanded and shown in blue lines. The black points are the waypoints of the yellow trajectories. The brown circle represents the gate point which is towards the grey box; (b) The evolution of the shape described by Fourier-based Representation; (c) The evolutionary changes in distance between the centroid of the particle group with the gate and the changes of the group size; (d) The evolution of the number of particles remains on the table and the pushing efficiency of a push; (e) The evolution of the group cohesion. (i) 74 particles; (ii) 95 particles; (iii) 128 particles; (iv) 140 candies.}
    \label{fig:exp_flow}
\end{figure*}

\section{Results} \label{sec:results}
The performance of the proposed method is evaluated through a series of experiments using a pair of UR-3 robots and two T-shaped tools. The robot observes the scene from the top through a RealSense camera D415. Data is passed to a Linux-based computer (i.e. Ubuntu 20) with the Robot Operating System (ROS) for image processing, decision-making, and robot control. We use GPT-4o as the task planner in the experiment.
We evaluate the performance of our method across several aspects: (1) shape representation, (2) path planning, and (3) cohesiveness of the particle group shape. 

\begin{figure}[t]
    \centering
    \includegraphics[width=1\linewidth]{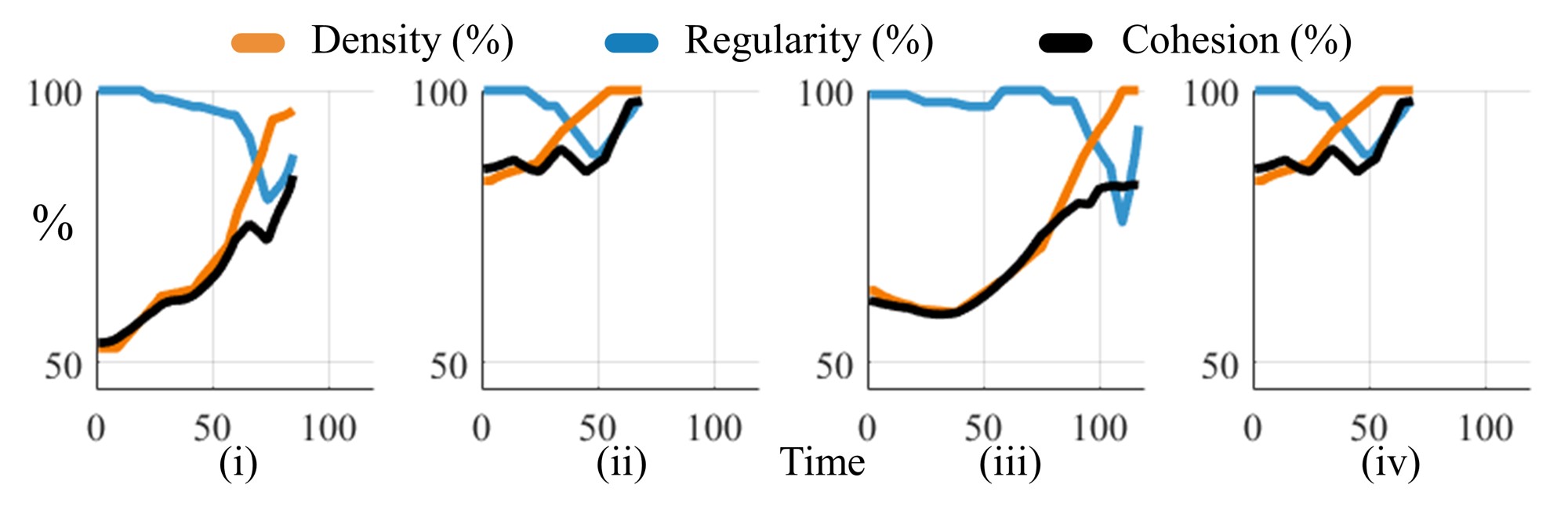}
    \caption{Evolution of the cohesion: (i)--(iv) reflect the changes in density, regularity, and cohesiveness among the particle groups in Fig. \ref{fig:exp_flow}(i)--(iv) cases.}
    \label{fig:cohesion}
\end{figure}

\begin{figure}[t]
    \centering
    \includegraphics[width=1\linewidth]{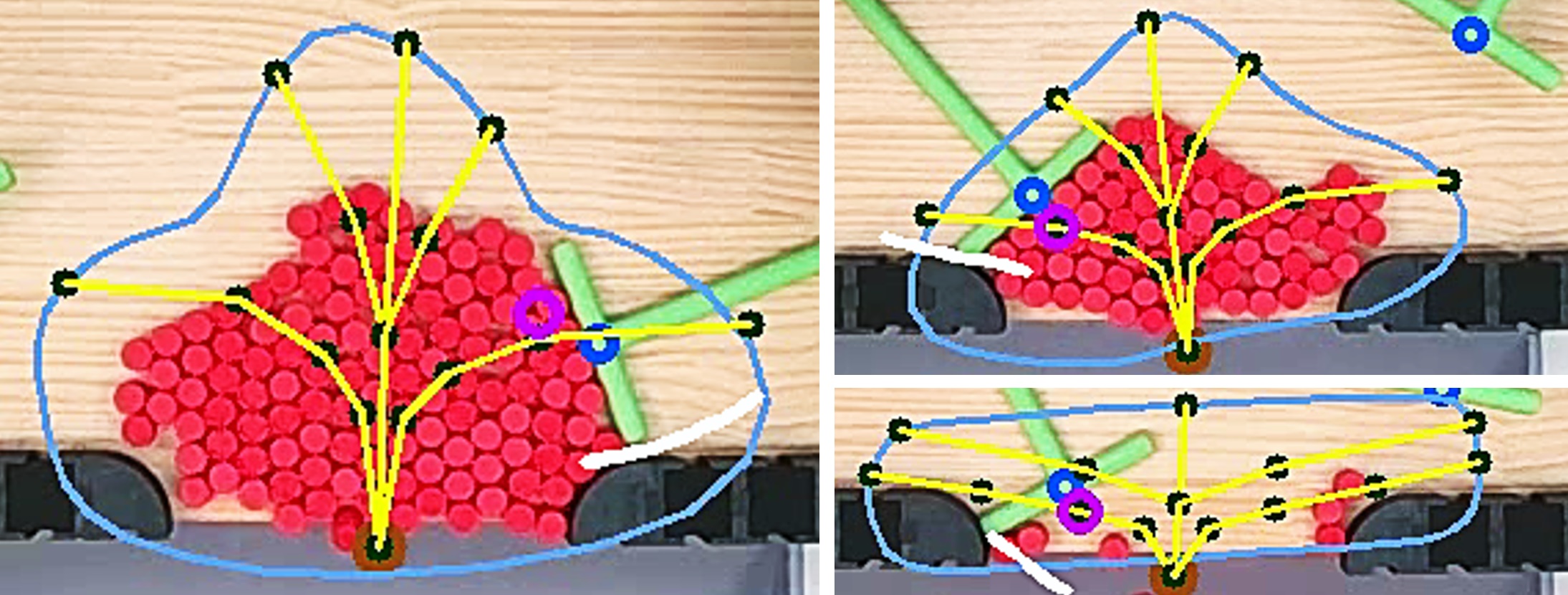}
    \caption{The white line represents the boundary calculated by the controller to avoid the tooltip from colliding with the wall during the manipulation process.}
    \label{fig:mpcExp}
\end{figure}

\subsection{Evaluation and Analysis}
To validate the robustness of our proposed model-free shape representation, we tested various particle group sizes and shapes, ranging from small (74 particles) to large (140 particles), as shown in Fig. \ref{fig:exp_flow}.

The robot first observes the scene, and the VLM provides a textual description with the information in the knowledge graph for the LLM to generate the next action. In the experiment, the robot first grasps the tool, then iteratively calculates the pushing direction. Through a series of guiding and pushing steps, the entire particle group is manipulated and guided into the designated grey container.

We begin by expanding the outline of the particle group (represented by the blue contour in Fig.\ref{fig:exp_flow}(a)). We then apply the Fourier series method described in Sec. \ref{sec:fourier} to this contour to represent the shape of the particle group. 
To strike a balance between shape representation fidelity and computational complexity, we set the number of harmonics $N$ to 5 and the number of farthest points concerned to 5.

Our experiments demonstrate that our system can successfully generate a Fourier-based shape representation to capture the arbitrary shape of the particle group in various circumstances. Fig. \ref{fig:exp_flow} illustrates the evolution of the particle group throughout the experiment, showing how the initially large, arbitrary shape is gradually moulded and compacted into a smaller form.

\begin{table*}[t]
\centering
\caption{Cohesion Comparison and Analysis}
\label{table:tbl_cohesionCompare} 
\begin{tabular}{c|ccc|cc|c|cccc}
\toprule
\multirow{3}{*}{Cases} & \multicolumn{6}{c|}{Regular Shapes} & \multicolumn{4}{c}{Experiments} \\ 
                       & \multicolumn{3}{c|}{Circle} & \multicolumn{2}{c|}{Square} & Rectangle & \multirow{2}{*}{Ours} & \multirow{2}{*}{MPC \cite{gold2022model}}  & \multirow{2}{*}{Landmark \cite{vardy2019landmark}}  & \multirow{2}{*}{Manual} \\ 
                       & $r = 2$ & $r = 4$ & $r = 8$ & $4\times4$ & $8\times8$ & $8\times2$ & &  & &   \\ \midrule
Density Ratio          & 0.5 & 0.5 & 0.5 & 0.5   & 0.5   & 0.5   & 0.846 & 0.917 & 0.698 & 0.866   \\ 
Regularity           & 1.0 & 1.0 & 1.0 & 0.934 & 0.934 & 0.682 & 0.811 & 0.685 & 0.828 & 0.809    \\ 
Cohesiveness           & 50.0\% & 50.0\% & 50.0\%& 46.7\%& 46.7\%& 34.1\%& 68.5\%& 62.4\%& 57.8\%& 70.1\% \\ \bottomrule
\end{tabular}
\end{table*}

During the manipulation process, the particles always remain cohesive, staying together and connected with one another.
To quantify the cohesion of the particle group, we compare the cohesiveness metrics calculated using Eq. \eqref{eq:cohesion}. As shown in Fig. \ref{fig:exp_flow}, the cohesiveness of the particle shape is preserved as the robot guides it to the desired location using the herding manipulation approach.
The evolution of the cohesiveness, from an initially low rate to the final high rate, is reflected in the changes in particle density and regularity, as plotted in Fig. \ref{fig:cohesion}. The experiments also demonstrate that when the particle density is high, the regularity tends to be at a lower value. This is primarily due to the iterative shaping of the particle group, which aims to keep it tightly packed while moving towards the gate. Importantly, the particle group is maintained as a single, cohesive entity, rather than dividing into multiple clusters or segmenting with large spaces. This proves the effectiveness of our proposed system in molding and herding the particle group as a whole.

With the Fourier-based shape descriptor, the system computes a tree-like trajectory, as shown by the yellow path with black waypoints in Fig. \ref{fig:exp_flow}. Given the arbitrary initial shape of the particle group, this demonstrates the system's capability to generate a reasonable trajectory for macro-scale shape manipulation.

The trajectory is further refined using the controller proposed in Sec. \ref{sec:mpc}. 
The prediction horizon is set to be 50. 
Fig. \ref{fig:mpcExp} shows the tool moving from a point $\mathbf{P}$ to the next waypoint, with the white line indicating the boundary for the tooltip movement based on the trajectory. The experiments with various particle shapes and group sizes demonstrate that the tool is able to push the particle group by following the trajectory of the action tree while keeping the tooltip from colliding with the walls.

\begin{figure}[t]
    \centering
    \includegraphics[width=1\linewidth]{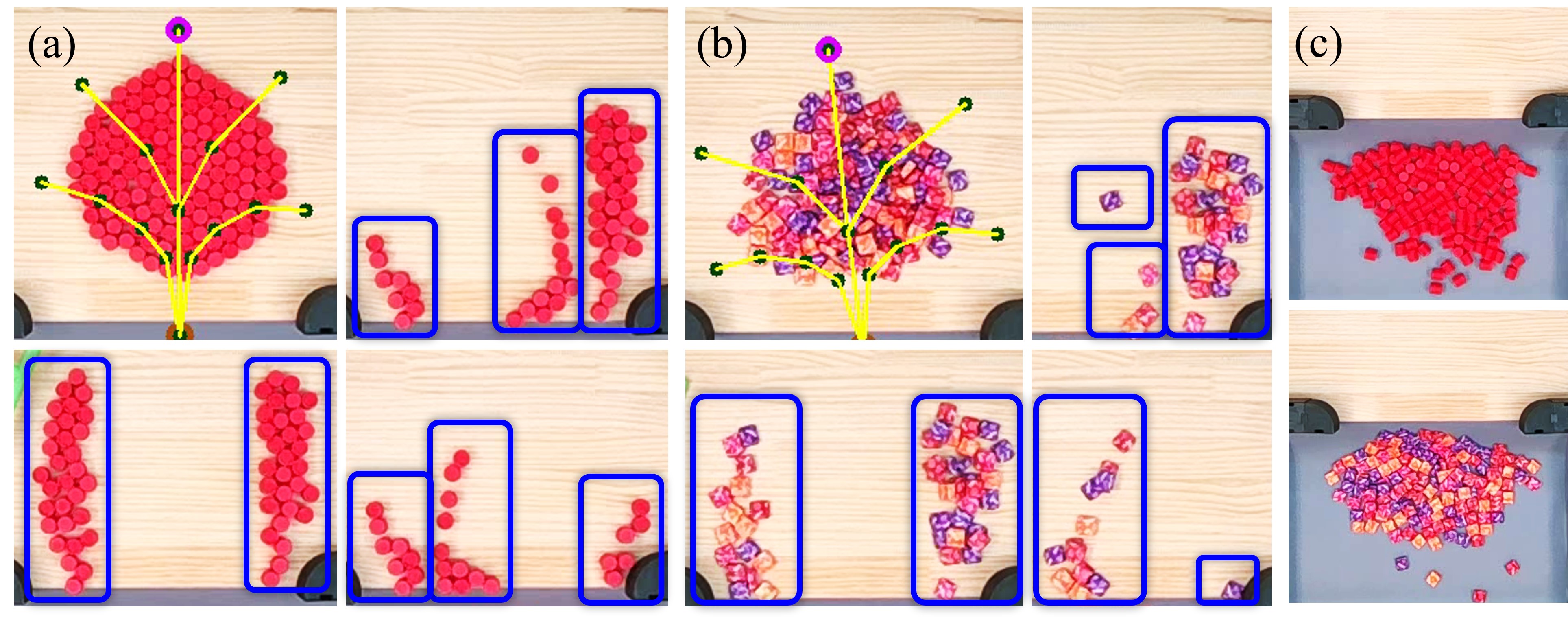}
    \caption{Performance of adopting direct pushing method: (a) 134 particles; (b) 140 candies; (c) All particles are successfully pushed into the container.}
    \label{fig:direct}
\end{figure}

\subsection{Comparison}
To demonstrate the advantage of the herding method over the direct brute force pushing approach, we conducted a comparative evaluation of the particle manipulation performance in terms of cohesiveness preservation.

We applied the same Fourier-based action tree strategy to derive the trajectory and executed it using the direct pushing technique. As shown in Fig. \ref{fig:direct}, the direct pushing approach failed to preserve the cohesion of the particle group, resulting in the division of the aggregate into multiple smaller subgroups. In contrast, our proposed shape manipulation action tree has been proven effective in handling scenarios involving multiple subgroups, successfully guiding all particles into the target box, as illustrated in the same figure.

The key distinction between the two methods lies in their underlying strategies. Our herding-based approach focuses on guiding the particles as a cohesive group, whereas the direct pushing technique tends to scatter the particles, leading to a loss of group cohesiveness.

To further quantify the cohesion properties, we present the cohesiveness metric results in Table \ref{table:tbl_cohesionCompare}. We first establish a baseline by comparing the cohesiveness of regular geometric shapes, including circles of varying radii, squares, and rectangles, all with a density ratio of 0.5. The results demonstrate that even when the density is the same, the cohesion can differ based on the shape. Specifically, circles exhibit the highest cohesion, followed by squares, while rectangles show the lowest cohesion due to the increased distance between the particles and the centroid.

We then apply our proposed cohesiveness metric to evaluate the performance of our system against state-of-the-art manipulation methods, including the MPC approach \cite{gold2022model} and a landmark-inspired technique \cite{vardy2019landmark}. The results indicate that while the traditional MPC method can achieve the highest density ratio, it exhibits the lowest cohesion among the evaluated approaches. This is primarily due to its focus on efficiently aggregating the particles rather than maintaining an optimal shape.

In contrast, the landmark-inspired method achieves the best regularity score, producing shapes that are close to the optimal configuration. This can be attributed to its shape-molding capabilities. Our proposed approach, on the other hand, generally demonstrates a better performance in preserving group cohesion, with the averagely high density and regularity values. However, when compared to manual human aggregation, our robotic system is still outperformed by approximately 1.5\% in terms of cohesiveness.

We illustrate the performance of the proposed methods in the accompanying video \url{https://vimeo.com/1043879712}.

\section{Conclusion} \label{sec:conclusions}
In this work, we have introduced a new Fourier-based shape control method and an iterative action tree for guiding and manipulating multiple particles in an aggregation task. To quantify the effectiveness of the proposed methodology, we have developed a cohesiveness metric to measure the compactness of a particle group. We have implemented the system with the ODKG and VLM for task planning and validated it using a dual-arm robotic platform. The experimental results show that our methodology achieves a cohesiveness of 68\%, while the human performance is slightly higher at 70\%. Although our system did not outperform the humans, these results are promising and indicate the potential of our approach. Moving forward, we plan to continue improving the cohesion preservation capabilities of our system. Future research directions may include enhancing the cohesiveness metric, exploring advanced control strategies and optimization techniques, investigating the scalability to handle larger systems, and extending the system to dynamic environments. By addressing these aspects, we aim to further advance the state-of-the-art in multi-object aggregation and manipulation, with the goal of developing intelligent robotic systems capable of efficient and cohesive group management in a wide range of applications.

\bibliography{david_biblio.bib}
\bibliographystyle{IEEEtran}

\vfill

\end{document}